\documentclass[runningheads]{llncs}
\usepackage[T1]{fontenc}
\usepackage{graphicx}
\usepackage{amsfonts}
\usepackage{amsmath}
\usepackage{tikz}
\usepackage{xcolor}
\usepackage{float}
\usepackage{makecell}
\usepackage{multirow}

\usepackage{microtype}
\usetikzlibrary{positioning,calc,arrows.meta,fit}
\usepackage[colorlinks=true,linkcolor=blue,citecolor=blue,urlcolor=blue]{hyperref}
\graphicspath{{figures/}}

\begin{document}
\title{Noise-Aware Visual Representation Learning for Medical Visual Question Answering}
%
%
\titlerunning{Noise-Aware Visual Representation Learning for Med-VQA}

\author{I Putu Adi Pratama \and Bahadorreza Ofoghi \and 
Atul Sajjanhar \and Shang Gao}
\authorrunning{I P. A. Pratama et al.}
%
\institute{Deakin University, 221 Burwood Highway, Burwood, Victoria 3125, Australia\\
\email{\{s225507154, b.ofoghi, atul.sajjanhar, shang.gao\}@deakin.edu.au}
}

\maketitle              
\begin{abstract}
Medical visual question answering (Med-VQA) has strong potential for clinical decision support by enabling AI models to interpret medical images and answer clinically relevant queries. Recent approaches typically connect off-the-shelf vision encoders with large language models (LLMs) through lightweight mapping networks to reduce computational cost. However, these methods often overlook the importance of handling noise and small irrelevant changes in visual representations. To address these challenges, we propose a noise-aware Med-VQA framework that incorporates a denoising autoencoder before visual embeddings are mapped into the input space of an LLM. The denoising autoencoder is pretrained to reconstruct clean visual embeddings from corrupted inputs, encouraging the model to learn robust visual representations that are less sensitive to noise. The resulting embeddings are then projected into the language model embedding space using a multi-layer perceptron (MLP), forming visual prefix tokens that provide image information to the LLM. To enable efficient adaptation without full retraining, we employ parameter-efficient fine-tuning using low-rank adaptation (LoRA). The proposed method is evaluated on the SLAKE and PathVQA benchmarks. Experimental results show improved robustness to noisy input embeddings while maintaining competitive clean performance across multiple evaluation criteria. These findings suggest that learning more robust visual representations can enhance Med-VQA performance and robustness. 

\keywords{Medical Visual Question Answering \and Vision-Language Models \and Visual Representation Learning \and Denoising Autoencoder}
\end{abstract}
%
%
%


\section{Introduction}
Medical visual question answering (Med-VQA) holds tremendous potential for clinical decision support by enabling artificial intelligence systems to interpret medical imagery and respond to clinical queries \cite{9434010,REZAEI2026111397}. Recent approaches typically leverage off-the-shelf frozen vision encoders and LLMs connected by mapping networks to reduce training costs. To ensure compatibility between visual features and the language model input space, various works employ specialized adapters to project visual representations into the LLM embedding space \cite{app15062983}. Several approaches utilize lightweight linear layers \cite{he2025pefomedparameterefficientfinetuning,li2023llavamed} or multi-layer perceptrons (MLPs) \cite{chen-etal-2024-towards-injecting,jiang2024medmoe,10.1007/978-3-031-43904-9_70,Zhang2024} to directly map visual embeddings into a sequence of learnable tokens for the LLM. For more advanced alignment, other works leverage transformer-based architectures, using sets of learnable queries and attention mechanisms to extract and align targeted visual features before projection \cite{10.1007/978-981-96-0908-6_6,ha-etal-2024-fusion,naseem2024biomedblip}.

Despite the cross-modal alignment capabilities of these adapter-based frameworks, existing approaches often do not explicitly address noisy or irrelevant variations in visual representations produced by pretrained vision encoders. Medical images are often affected by various types and levels of noise introduced during acquisition and transmission \cite{Jifara2019,7836672}, which can complicate diagnosis and downstream analysis. As a result, projection-based adapters may propagate both clinically relevant and noisy visual information into the LLM. This motivates the need for representation learning approaches that improve the robustness of visual embeddings before they are mapped into the LLM embedding space, an area that remains relatively underexplored in Med-VQA. In this context, denoising autoencoder paradigms \cite{10.1145/1390156.1390294,10.5555/1756006.1953039} provide a suitable strategy by learning representations that are robust to input corruption. In this work, Gaussian noise is used as the corruption model because visual embeddings produced by the pretrained vision encoder are continuous real-valued vectors, making additive Gaussian perturbation a simple, controlled, and reproducible way to simulate small variations in the embedding space \cite{10.5555/1756006.1953039}. By reconstructing clean visual embeddings from Gaussian-corrupted inputs, the model is encouraged to capture stable underlying structures while suppressing noise-related variations before projection into the LLM embedding space \cite{DBLP:journals/corr/abs-2201-03898,10.1145/1390156.1390294}.

\begin{figure}[t]
    \centering
    \includegraphics[width=\linewidth]{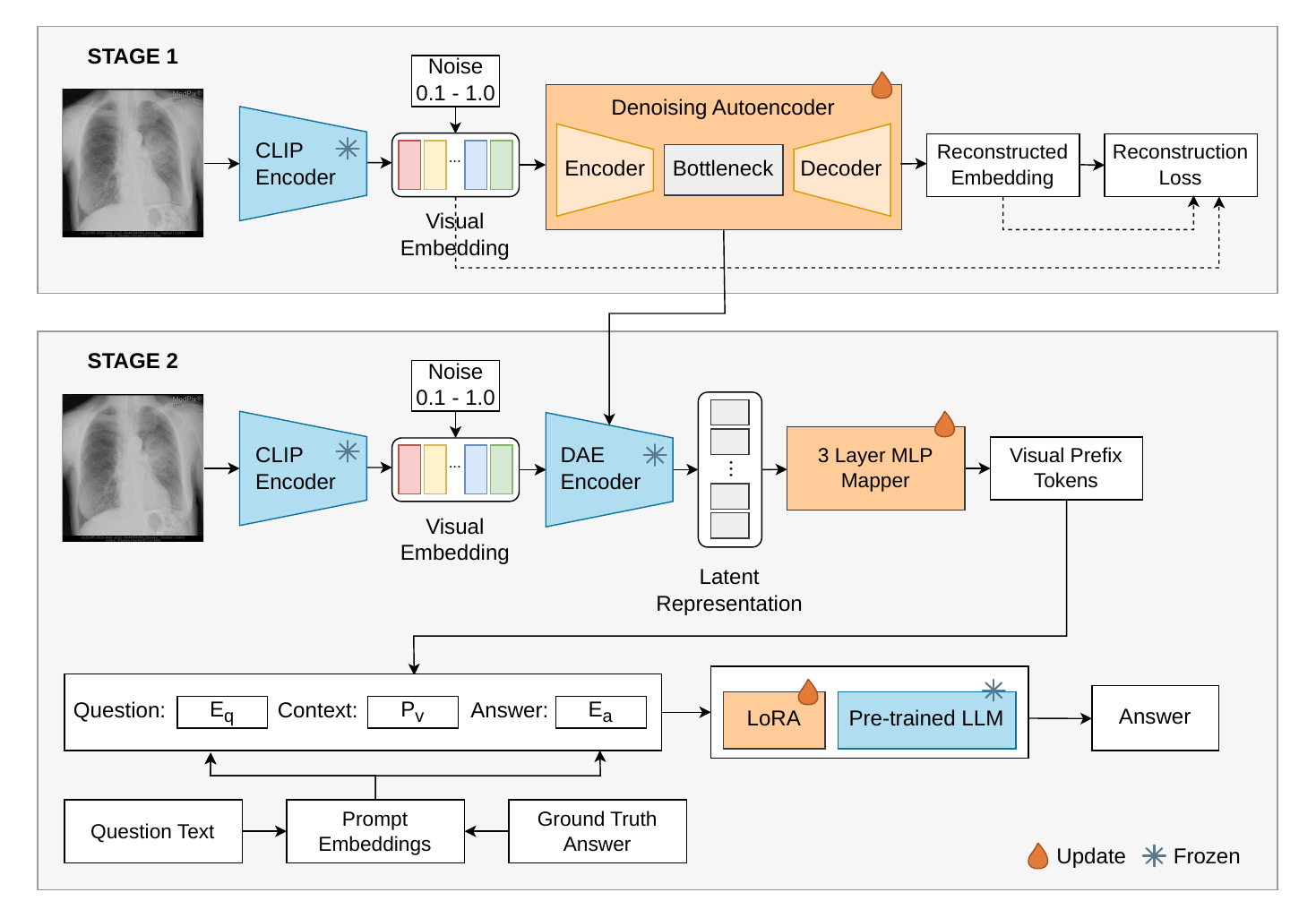}
    \caption{Overview of the proposed two-stage Med-VQA framework. Stage 1 trains a denoising autoencoder on visual embeddings using a reconstruction objective to learn robust latent representations from corrupted inputs. Stage 2 reuses the frozen denoising autoencoder encoder (DAE Encoder), projects the latent representation through a 3-layer MLP mapper into visual prefix tokens, and conditions a frozen LLM, optionally adapted with LoRA, on the visual prefix together with textual embeddings.}
    \label{fig:fig1}
\end{figure}

In this work, we propose a noise-aware visual representation learning framework for Med-VQA, as illustrated in Figure~\ref{fig:fig1}. The framework incorporates a denoising autoencoder before the 3-layer MLP mapper to learn robust, noise-aware representations by reconstructing clean visual embeddings from corrupted embeddings produced by a frozen CLIP encoder. The resulting representations are then projected into the embedding space of a frozen LLM. In the first stage, we focus on visual representation learning, where visual embeddings extracted from the frozen vision encoder are explicitly corrupted and passed through a denoising autoencoder with a dimensional bottleneck. The autoencoder is trained with a reconstruction objective to learn robust, noise-aware representations.

In the second stage, the learned latent representations are integrated into the generative Med-VQA pipeline. Specifically, these embeddings are projected via an MLP mapper into visual prefix tokens, which are then used to prompt the LLM for answer generation. This stage is optimized with a language modeling objective, enabling effective cross-modal alignment between visual and textual modalities. Furthermore, to adapt the LLM to the medical domain without catastrophic forgetting, we employ Low-Rank Adaptation (LoRA)-based parameter-efficient fine-tuning \cite{hu2022lora}, updating only a small subset of attention parameters while keeping the backbone models frozen.

Experimental results across the SLAKE and PathVQA benchmarks show that this two-stage design yields consistent robustness gains under noisy evaluation. On SLAKE, our proposed denoising autoencoder framework improves performance under noisy evaluation conditions over the baseline in both LoRA and frozen settings, with the average accuracy across noisy evaluation settings increasing from \(0.642\) to \(0.735\) for LoRA and from \(0.473\) to \(0.713\) for the frozen setting. On PathVQA, the proposed framework shows the same trend, with the average accuracy across noisy evaluation settings increasing from \(0.514\) to \(0.568\) for LoRA and from \(0.348\) to \(0.518\) for the frozen setting. Compared with projection-only baselines that directly map visual embeddings produced by the pretrained vision encoder into the LLM embedding space, as well as standard autoencoder variants, the proposed denoising autoencoder framework is more resilient to corrupted input embeddings. Overall, reconstructing clean visual embeddings from corrupted embeddings before mapping them into the LLM embedding space helps the model learn more robust representations, improving the stability and effectiveness of Med-VQA answer generation. These findings suggest that embedding-level denoising is a promising strategy for improving robustness in Med-VQA systems trained on relatively small medical datasets.


To summarize, our main contributions are as follows:
\begin{itemize}
    \item \textbf{Two-Stage Noise-Aware Visual Representation Learning for Med-VQA:} We propose a two-stage training paradigm that decouples denoising representation learning from generative alignment in Med-VQA. In the first stage, a denoising autoencoder is trained to reconstruct clean visual embeddings from noise-corrupted inputs, encouraging the model to learn representations that are robust to irrelevant perturbations. In the second stage, these robust visual representations are projected into the language model embedding space for answer generation.
        
    \item \textbf{Robustness Evaluation under Noisy Conditions:} We propose a structured robustness evaluation for generative Med-VQA by systematically injecting noise into visual embeddings produced by the pretrained vision encoder during inference. This provides a controlled and direct assessment of how direct projection, autoencoder-based representation learning, and denoising autoencoder-based representation learning perform under clean and corrupted embedding conditions.
\end{itemize}

\section{Related Work}
Generative medical VQA approaches usually adopt a modular design, combining frozen vision encoders and LLMs through learnable mapping networks. A representative framework maps visual features extracted from a CLIP vision encoder into a sequence of prefix tokens via an MLP, which are then used to condition pretrained language models \cite{10.1007/978-3-031-43904-9_70}. In a similar direction, domain-adapted vision--language models integrate biomedical vision encoders with radiology-specific LLMs using query transformers and MLP-based projections \cite{ha-etal-2024-fusion}. 

Extending this paradigm, subsequent works incorporate 3D medical image encoding and Q-Former-style query modules to map visual features into visual prefix tokens, followed by lightweight projection mechanisms to align visual and textual representations \cite{10.1007/978-981-96-0908-6_6}. Likewise, BLIP-style architectures employ learnable transformation layers to project visual features into the language embedding space, improving cross-modal alignment \cite{naseem2024biomedblip}. More recent methods further connect a CLIP vision encoder with LLaMA-based LLMs through lightweight projection modules, enabling multimodal instruction-following and open-ended VQA \cite{li2023llavamed}. In addition, mixture-of-experts frameworks such as Med-MoE first use an MLP after the vision encoder to align medical image tokens with language tokens, before introducing routing mechanisms that selectively activate domain-specific experts for multimodal reasoning \cite{jiang2024medmoe}.

While these approaches have demonstrated the effectiveness of multimodal alignment and generative adaptation, they generally assume that the visual embeddings produced by pretrained vision encoders are sufficiently informative and reliable. In practice, however, such embeddings may contain noisy, redundant, or task-irrelevant variations inherited from large-scale pretraining, which can be propagated into the language model through the projection module. Consequently, improving the quality of visual representations before multimodal alignment may further enhance downstream reasoning performance. This motivates the use of denoising representation learning, where the model learns robust visual representations by reconstructing clean embeddings from corrupted inputs \cite{10.1145/1390156.1390294,10.5555/1756006.1953039}. 

Denoising autoencoders provide a principled approach for learning representations that are robust to irrelevant perturbations, and prior work has shown that denoising objectives can improve representation robustness and downstream learning performance \cite{10.1145/1390156.1390294,10.5555/1756006.1953039,10.5555/2627435.2750359}. In medical imaging, denoising approaches have been applied to address acquisition-related noise and improve downstream analysis in tasks such as image reconstruction, clinical detection, and anomaly detection \cite{7836672,KASCENAS2023102963,rahman2023taskspecific}. However, most existing studies focus on denoising image inputs or intermediate feature representations. The use of denoising representation learning to refine visual embeddings produced by the pretrained vision encoder prior to their projection into the LLM embedding space remains relatively underexplored in Med-VQA systems. 

\section{Methodology}

\subsection{Model Architecture}
\label{subsec:model_architecture}

\label{subsec:model_architecture}

Our noise-aware Med-VQA framework consists of four main components: a frozen CLIP encoder, a denoising autoencoder, a 3-layer MLP mapper, and a pretrained LLM, as illustrated in Fig.~\ref{fig:fig1}. The framework follows a two-stage design that first learns robust visual representations and then aligns them with the language model for answer generation.

\paragraph{Visual representation.}
Given an image $I$, a frozen CLIP encoder \cite{pmlr-v139-radford21a} extracts a global visual embedding $\mathbf{v} \in \mathbb{R}^{d_v}$.

\paragraph{Denoising autoencoder.}
The visual embedding $\mathbf{v}$ is processed by a denoising autoencoder to obtain a latent representation. During denoising autoencoder pretraining, a corrupted embedding $\tilde{\mathbf{v}}$ is constructed by adding noise to $\mathbf{v}$. The encoder maps the corrupted embedding into a latent representation $\mathbf{z}=f_{\text{enc}}(\tilde{\mathbf{v}})$, and the decoder reconstructs the original clean embedding as $\hat{\mathbf{v}}=f_{\text{dec}}(\mathbf{z})$. After pretraining, only the encoder is retained and used to produce robust visual representations for Med-VQA training.

\paragraph{Visual-prefix mapping.}
The latent representation $\mathbf{z}$ produced by the pretrained denoising autoencoder's encoder is projected into the language model embedding space using a 3-layer MLP, where $\mathbf{p} = f_{\text{map}}(\mathbf{z})$. The output $\mathbf{p} \in \mathbb{R}^{L_v d_h}$ is reshaped into visual prefix tokens $\mathbf{P}_v \in \mathbb{R}^{L_v \times d_h}$, where $L_v$ denotes the number of visual tokens and $d_h$ is the hidden dimension of the language model.

\paragraph{Text conditioning and decoding.}
Given a question \(q\) and answer \(a\), we construct textual scaffolds \(\texttt{"question: "}\), \(\texttt{" context:"}\), and \(\texttt{"answer "}\). Let \(\mathbf{E}_{q}\) and \(\mathbf{E}_{a}\) denote the token embeddings of the question and answer text, and let \(\mathbf{P}_{v}\) be the projected visual-prefix embeddings. The input sequence is formed as
\[
\mathbf{X}=
[\mathbf{E}_{\texttt{question: }},
\mathbf{E}_{q},
\mathbf{E}_{\texttt{ context:}},
\mathbf{P}_{v},
\mathbf{E}_{\texttt{answer }},
\mathbf{E}_{a}],
\]
where \(\mathbf{E}_{a}\) is included during training, while at inference the model starts from \(\texttt{"answer "}\) and generates tokens autoregressively. A causal language model then generates the answer conditioned on both the visual-prefix tokens \(\mathbf{P}_{v}\) and the textual context tokens.

\subsection{Training Approach}

The framework is trained in two stages: (1) denoising autoencoder pretraining, and (2) task-specific Med-VQA training. This separation allows the model to first learn robust visual representations from corrupted inputs before adapting them for downstream answer generation.

\paragraph{Stage 1: Denoising autoencoder pretraining.}
In the first stage, the denoising autoencoder is pretrained independently using an embedding reconstruction objective. Given a clean visual embedding $\mathbf{v}$, Gaussian noise is injected to obtain a corrupted embedding $\tilde{\mathbf{v}} = \mathbf{v} + \epsilon$, where $\epsilon \sim \mathcal{N}(0, \sigma^2 I)$. The corrupted embedding is encoded into a latent representation $\mathbf{z} = f_{\text{enc}}(\tilde{\mathbf{v}})$ and decoded to reconstruct the original clean embedding $\hat{\mathbf{v}} = f_{\text{dec}}(\mathbf{z})$.

The denoising autoencoder is trained using a Smooth L1 reconstruction objective:
\[
\mathcal{L}_{\text{DAE}} =
\frac{1}{N} \sum_{i=1}^{N}
\begin{cases}
\frac{1}{2} (\hat{v}_i - v_i)^2 & \text{if } |\hat{v}_i - v_i| < 1 \\
|\hat{v}_i - v_i| - \frac{1}{2} & \text{otherwise}.
\end{cases}
\]

In this equation, \(\mathcal{L}_{\text{DAE}}\) denotes the reconstruction loss of the denoising autoencoder, \(N\) is the total number of embedding dimensions, \(v_i\) represents the original clean visual embedding value at dimension \(i\), and \(\hat{v}_i\) represents the reconstructed embedding value produced by the denoising autoencoder.

By reconstructing clean embeddings from corrupted inputs, the denoising autoencoder encourages the encoder to learn visual representations that are more robust to noisy perturbations. After pretraining, the decoder is discarded and only the encoder is used in Stage 2.

\paragraph{Stage 2: Med-VQA training.}
In the second stage, the pretrained denoising autoencoder encoder is integrated into the Med-VQA pipeline. For each input image $I$, the visual embedding $\mathbf{v}$ is transformed into a latent representation $\mathbf{z} = f_{\text{enc}}(\mathbf{v})$, which is then mapped into visual prefix tokens $\mathbf{P}_v = f_{\text{map}}(\mathbf{z})$.

The language model is trained to generate the answer sequence conditioned on both visual and textual inputs using the objective:
\[
\mathcal{L}_{\text{VQA}} = -\sum_{t \in \mathcal{A}} \log p_\theta(y_t \mid y_{<t}, q, I),
\]
where \(I\) denotes the input image, \(q\) denotes the question, \(y_t\) denotes the target answer token at position \(t\), \(y_{<t}\) represents the previously generated answer tokens before position \(t\), \(p_\theta\) denotes the probability distribution parameterised by the model parameters \(\theta\), and \(\mathcal{A}\) denotes the set of answer-token positions used in the loss calculation.

During Med-VQA training, the vision encoder and the pretrained denoising autoencoder's encoder remain frozen. Depending on the experimental setting, either only the visual mapper is trained or parameter-efficient fine-tuning using LoRA is applied to adapt the language model. Compared with the fully frozen setting, LoRA enables parameter-efficient task-specific adaptation within the language model while keeping the pretrained backbone fixed, allowing answer generation to better incorporate the visual prefix tokens and question context.

\section{Experimental Setup}

\paragraph{\textbf{Experimental Objectives.}} 
We investigate the effectiveness of applying a denoising autoencoder to CLIP visual embeddings for learning more robust visual representations for Med-VQA. These representations are evaluated based on their downstream question-answering performance and robustness to noisy perturbations. We assess these properties through two groups of experiments: (1) downstream Med-VQA performance comparison, including ablation comparisons among the baseline, AE, and DAE variants, and (2) robustness evaluation under noisy embedding conditions.

\paragraph{\textbf{Datasets.}}
We evaluate the proposed method on two publicly available Med-VQA datasets: \textbf{SLAKE} \cite{9434010} and \textbf{PathVQA} \cite{he-etal-2021-towards}. These datasets cover diverse imaging modalities and question types, including both open-ended and yes/no formats. SLAKE provides well-annotated radiology images for structured evaluation, and PathVQA focuses on histopathology images requiring fine-grained visual understanding. Together, they enable a comprehensive evaluation of robustness and generalisation. We use the official train, validation, and test splits for all datasets.

\begin{table}[t]
\centering
\caption{Statistics of the Med-VQA datasets used in this paper.}
\label{tab:dataset_stats}
\begin{tabular}{lcc}
\hline
 & SLAKE & PathVQA \\
\hline
Number of images & 642 & 4,998 \\
Number of questions & 14,028 & 32,799 \\
Number of unique answers & 461 & 3,182 \\
\hline
\end{tabular}
\end{table}

\paragraph{\textbf{Evaluation Metrics.}}
We evaluate the proposed method using standard metrics for generative visual question answering, including BLEU, BERTScore, F1 score, and Accuracy. Accuracy is reported as overall accuracy as well as per question type, including \textit{open-ended} and \textit{Yes/No} questions.

\paragraph{\textbf{Implementation Details.}}
We use precomputed visual embeddings extracted from a frozen CLIP ViT-B/32 vision encoder and train a generative Med-VQA model with a frozen or parameter-efficiently tuned GPT-2 XL language model. Specifically, each input image is resized to $224 \times 224$ and processed by the CLIP encoder, which divides the image into $32 \times 32$ patches, resulting in a $7 \times 7$ grid (49 patches). The encoder produces a global visual embedding of dimension 512, which is used as the input representation.

The denoising autoencoder is first pretrained on an embedding reconstruction task, where noise is injected into visual embeddings and the model learns to reconstruct the corresponding clean embeddings. In our implementation, the visual embedding input dimension is $512$, and noise is added as $\mathbf{x}_{\text{noisy}}=\mathbf{x}+\boldsymbol{\epsilon}$ with $\boldsymbol{\epsilon}\sim\mathcal{N}(0,\sigma^2\mathbf{I})$. We evaluate three embedding corruption levels, $\sigma \in \{0.1, 0.5, 1.0\}$, to represent mild, moderate, and severe perturbations relative to the empirical spread of the CLIP embedding space. This provides a compact, data-scaled, and computationally practical protocol for analysing the effect of different corruption strengths during Stage-1 representation learning. Gaussian corruption is used as a simple and controlled perturbation strategy, allowing each embedding dimension to be perturbed in a systematic and reproducible manner under different noise levels \cite{Jifara2019}.

The denoising autoencoder employs a bottleneck MLP architecture that compresses 512-dimensional embeddings through a 256-dimensional hidden layer into a 128-dimensional latent representation, before reconstructing them back to 512 dimensions in the decoder. ReLU activations are applied between layers, and input dropout with a rate of 0.1 is used during training. Reconstruction is optimised against the clean embedding using Smooth L1 loss. This pretraining encourages the encoder to learn robust visual representations that are less sensitive to noisy perturbations.

These latent representations are subsequently projected into visual prefix tokens through an MLP mapper. Specifically, given a latent embedding $\mathbf{z} \in \mathbb{R}^{d_z}$, the mapping network $f_M$ transforms it into a sequence of prefix tokens that are compatible with the GPT-2 XL embedding space. 

The MLP mapper is implemented as a three-layer feed-forward network with dimensions $\{d_z, \frac{\ell_x \cdot e}{2}, \ell_x \cdot e\}$, where $\ell_x$ denotes the prefix length and $e$ is the embedding dimension of the language model. In our setting, the latent dimension is $d_z = 128$, the prefix length is set to $\ell_x = 8$, and the GPT-2 XL embedding size is $e = 1600$. The output of the MLP is reshaped into a sequence of $\ell_x$ visual prefix tokens, each with embedding dimension $e$, resulting in \(P_v \in \mathbb{R}^{\ell_x \times e}\), which is then inserted into the input sequence.

The question text is tokenized using the GPT-2 XL tokenizer and mapped through the GPT-2 XL token embedding layer. This combined sequence is then fed to GPT-2 XL, which generates the answer autoregressively, predicting each token conditioned on the question and visual prefix context. During training, the model is optimized with a language-modeling objective (cross-entropy) to maximize the likelihood of the ground-truth answer tokens.

We evaluate the framework under both fully frozen and LoRA-based parameter-efficient fine-tuning settings.

\section{Results and Discussion}

\subsection{Overall Performance}

We compare three settings: (1) a baseline Med-VQA model that directly maps visual embeddings to the language model \cite{10.1007/978-3-031-43904-9_70}, (2) an autoencoder-enhanced model without noise injection, and (3) the proposed denoising autoencoder model with noise injection. This comparison isolates the contributions of autoencoder-based representation learning and denoising-based representation learning to downstream Med-VQA performance.

Table~\ref{tab:main_results} presents the clean evaluation results on the SLAKE and PathVQA datasets under both Frozen and LoRA fine-tuning settings. Performance is evaluated using BLEU-1 (BL1), BERTScore (BS), F1-score, and accuracy. The results show that the effect of autoencoder-based representation learning differs across datasets and fine-tuning settings.

On SLAKE, the proposed DAE-based framework achieves the strongest overall performance. Under the LoRA setting, it obtains the best BLEU-1, BERTScore, and F1-score, with scores of 0.796, 0.923, and 0.810, respectively, while matching the highest accuracy of 0.756. Under the Frozen setting, the DAE-based framework also outperforms both the baseline and AE variants, achieving the highest BLEU-1, BERTScore, F1-score, and accuracy.

On PathVQA, the AE-based model with LoRA achieves the best clean performance, obtaining a BLEU-1 score of 0.620, BERTScore of 0.804, F1-score of 0.624, and accuracy of 0.602. Although the AE-based model achieves the highest clean accuracy on PathVQA, the DAE-based model demonstrates substantially better robustness under noisy embedding conditions, as shown in the following section.

Overall, these results indicate that learning more robust visual representations before projection into the LLM embedding space can benefit Med-VQA performance, particularly on SLAKE. However, the clean evaluation results also show that the DAE-based framework does not always achieve the highest clean accuracy across all datasets. This motivates the robustness analysis in the following section, where the effect of denoising representation learning is further evaluated under noisy embedding conditions.

\begin{table}[t]
\centering
\caption{Comparison on SLAKE and PathVQA using clean evaluation. Best results are shown in bold. For Ours w/ DAE, values are reported from the Stage-1 noise \(\sigma=0.5\) setting for consistency with the robustness analysis.}
\footnotesize
\setlength{\tabcolsep}{3pt}
\renewcommand{\arraystretch}{1.25}
\resizebox{\textwidth}{!}{
\begin{tabular}{l l cccc cccc}
\hline
 & LM fine-tuning &
\multicolumn{4}{c}{SLAKE} &
\multicolumn{4}{c}{PathVQA} \\
\cline{3-10}
Model & Setting &
BL1 & BS & F1 & Acc. &
BL1 & BS & F1 & Acc. \\
\hline

Baseline & Frozen 
& 0.756 & 0.908 & 0.764 & 0.719
& 0.553 & 0.777 & 0.560 & 0.540 \\[2pt]

Baseline & LoRA 
& 0.793 & 0.922 & 0.808 & 0.748
& 0.614 & 0.802 & 0.619 & 0.598 \\[2pt]
\hline

Ours w/ AE & Frozen 
& 0.767 & 0.914 & 0.747 & 0.700
& 0.555 & 0.778 & 0.561 & 0.544 \\[2pt]

Ours w/ AE & LoRA 
& 0.800 & 0.922 & 0.809 & \textbf{0.756}
& \textbf{0.620} & \textbf{0.804} & \textbf{0.624} & \textbf{0.602} \\[2pt]
\hline

Ours w/ DAE & Frozen 
& 0.776 & 0.916 & 0.786 & 0.743
& 0.553 & 0.775 & 0.556 & 0.543 \\[2pt]

Ours w/ DAE & LoRA 
& \textbf{0.796} & \textbf{0.923} & \textbf{0.810} & \textbf{0.756}
& 0.607 & 0.796 & 0.612 & 0.592 \\[2pt]
\hline

\end{tabular}
}
\label{tab:main_results}
\end{table}

\subsection{Robustness to Noisy Visual Embeddings}

The robustness evaluation is conducted by training the denoising autoencoder with Gaussian noise during Stage 1 and injecting Gaussian noise into the visual embeddings during inference. Specifically, the denoising autoencoder is trained with different Stage-1 noise levels \((\sigma \in \{0.1, 0.5, 1.0\})\), while inference-time noise is applied to compare the robustness of the baseline, AE-based, and proposed DAE-based frameworks under noisy conditions.

\begin{table}[!htbp]
\centering
\caption{SLAKE robustness evaluation under different Stage-1 corruption levels for LoRA and Frozen settings}
\label{tab:stage1_noise_combined}
\begin{tabular}{ccccccc}
\hline
Setting & Stage-1 Noise & Acc@0 & Acc@0.1 & Acc@0.5 & Acc@1.0 & Noisy Avg. Acc. \\
\hline
\multirow{3}{*}{LoRA} 
& 1.00 & 0.736 & 0.735 & 0.729 & 0.692 & 0.719 \\
& 0.50 & 0.756 & \textbf{0.756} & \textbf{0.746} & \textbf{0.703} & \textbf{0.735} \\
& 0.10 & \textbf{0.769} & 0.750 & 0.716 & 0.633 & 0.700 \\
\hline
\multirow{3}{*}{Frozen} 
& 1.00 & 0.700 & 0.700 & 0.701 & \textbf{0.673} & 0.691 \\
& 0.50 & \textbf{0.743} & \textbf{0.744} & \textbf{0.726} & 0.669 & \textbf{0.713} \\
& 0.10 & 0.714 & 0.713 & 0.658 & 0.508 & 0.626 \\
\hline
\end{tabular}
\end{table}

\begin{table}[!htbp]
\centering
\small
\caption{Comparison of robustness performance on SLAKE dataset under clean and noisy embedding conditions}
\label{tab:slake_noisy_avg_latest}
\begin{tabular}{l l c c c c c}
\hline
\multicolumn{1}{c}{Setting} & \multicolumn{1}{c}{Setup} & Acc@0 & Acc@0.1 & Acc@0.5 & Acc@1.0 & Noisy Avg. Acc. \\
\hline
\multirow{3}{*}{LoRA}
& Baseline    & 0.748 & 0.749 & 0.637 & 0.539 & 0.642 \\
& Ours w/ AE  & 0.756 & 0.751 & 0.668 & 0.563 & 0.661 \\
& Ours w/ DAE & \textbf{0.756} & \textbf{0.756} & \textbf{0.746} & \textbf{0.703} & \textbf{0.735} \\
\hline
\multirow{3}{*}{Frozen}
& Baseline    & 0.719 & 0.712 & 0.472 & 0.235 & 0.473 \\
& Ours w/ AE  & 0.700 & 0.697 & 0.555 & 0.432 & 0.561 \\
& Ours w/ DAE & \textbf{0.743} & \textbf{0.744} & \textbf{0.726} & \textbf{0.669} & \textbf{0.713} \\
\hline
\end{tabular}
\end{table}

\begin{table}[!htbp]
\centering
\caption{PathVQA robustness evaluation under different Stage-1 corruption levels for LoRA and Frozen settings}
\label{tab:pathvqa_stage1_noise_combined}
\begin{tabular}{ccccccc}
\hline
Setting & Stage-1 Noise & Acc@0 & Acc@0.1 & Acc@0.5 & Acc@1.0 & Noisy Avg. Acc. \\
\hline
\multirow{3}{*}{LoRA}
& 1.00 & 0.576 & 0.575 & 0.568 & 0.549 & 0.564 \\
& 0.50 & 0.592 & \textbf{0.592} & \textbf{0.576} & \textbf{0.535} & \textbf{0.568} \\
& 0.10 & \textbf{0.601} & 0.595 & 0.534 & 0.479 & 0.536 \\
\hline
\multirow{3}{*}{Frozen}
& 1.00 & 0.521 & 0.520 & 0.514 & 0.492 & 0.509 \\
& 0.50 & 0.543 & \textbf{0.544} & \textbf{0.527} & \textbf{0.483} & \textbf{0.518} \\
& 0.10 & \textbf{0.549} & 0.543 & 0.473 & 0.403 & 0.473 \\
\hline
\end{tabular}
\end{table}

\begin{table}[!htbp]
\centering
\small
\caption{Comparison of robustness performance on PathVQA dataset under clean and noisy embedding conditions}
\label{tab:pathvqa_noisy_avg_latest}
\begin{tabular}{l l c c c c c}
\hline
\multicolumn{1}{c}{Setting} & \multicolumn{1}{c}{Setup} & Acc@0 & Acc@0.1 & Acc@0.5 & Acc@1.0 & Noisy Avg. Acc. \\
\hline
\multirow{3}{*}{LoRA}
& Baseline    & 0.598 & 0.585 & 0.500 & 0.458 & 0.514 \\
& Ours w/ AE  & \textbf{0.602} & \textbf{0.597} & 0.521 & 0.482 & 0.533 \\
& Ours w/ DAE & 0.592 & 0.592 & \textbf{0.576} & \textbf{0.535} & \textbf{0.568} \\
\hline
\multirow{3}{*}{Frozen}
& Baseline    & 0.540 & 0.521 & 0.336 & 0.187 & 0.348 \\
& Ours w/ AE  & \textbf{0.544} & 0.531 & 0.429 & 0.370 & 0.443 \\
& Ours w/ DAE & 0.543 & \textbf{0.544} & \textbf{0.527} & \textbf{0.483} & \textbf{0.518} \\
\hline
\end{tabular}
\end{table}

Table~\ref{tab:stage1_noise_combined} and Table~\ref{tab:pathvqa_stage1_noise_combined} present the robustness results under different Stage-1 corruption levels on SLAKE and PathVQA, respectively. The Stage-1 Noise column denotes the standard deviation of Gaussian noise injected into the CLIP image embeddings during denoising autoencoder training. Acc@0 represents clean evaluation, while Acc@0.1, Acc@0.5, and Acc@1.0 report accuracy when Gaussian noise with standard deviations of 0.1, 0.5, and 1.0 is injected during inference. Noisy Avg. Acc. is computed as the average of Acc@0.1, Acc@0.5, and Acc@1.0.

Across both datasets, moderate Stage-1 corruption ($\sigma=0.50$) provides the best overall robustness. On SLAKE, it achieves the highest Noisy Avg. Acc. of 0.735 under LoRA and 0.713 under Frozen. Similarly, on PathVQA, it achieves the highest Noisy Avg. Acc. of 0.568 under LoRA and 0.518 under Frozen. In contrast, the lowest corruption level ($\sigma=0.10$) often gives the best clean accuracy, but its performance declines more noticeably as inference-time noise increases. This suggests that weak corruption may favour clean-condition performance, while moderate corruption provides a better trade-off between clean accuracy and robustness.

Table~\ref{tab:slake_noisy_avg_latest} and Table~\ref{tab:pathvqa_noisy_avg_latest} compare the baseline, AE, and DAE variants under clean and noisy embedding conditions. On SLAKE, the DAE-based model achieves the strongest robustness in both LoRA and Frozen settings. Under LoRA, Noisy Avg. Acc. improves from 0.642 for the baseline to 0.735 with DAE. Under Frozen, it improves from 0.473 to 0.713. The advantage is especially clear under severe noise, where DAE achieves 0.703 under LoRA and 0.669 under Frozen, clearly outperforming both the baseline and AE variants.

A similar trend is observed on PathVQA. Under LoRA, DAE improves Noisy Avg. Acc. from 0.514 for the baseline to 0.568. Under Frozen, the improvement is larger, increasing from 0.348 to 0.518. Although AE achieves slightly higher clean accuracy in some PathVQA settings, DAE consistently performs better under moderate and severe noise levels.

\begin{figure}[!htbp]
\centering
\includegraphics[width=\textwidth]{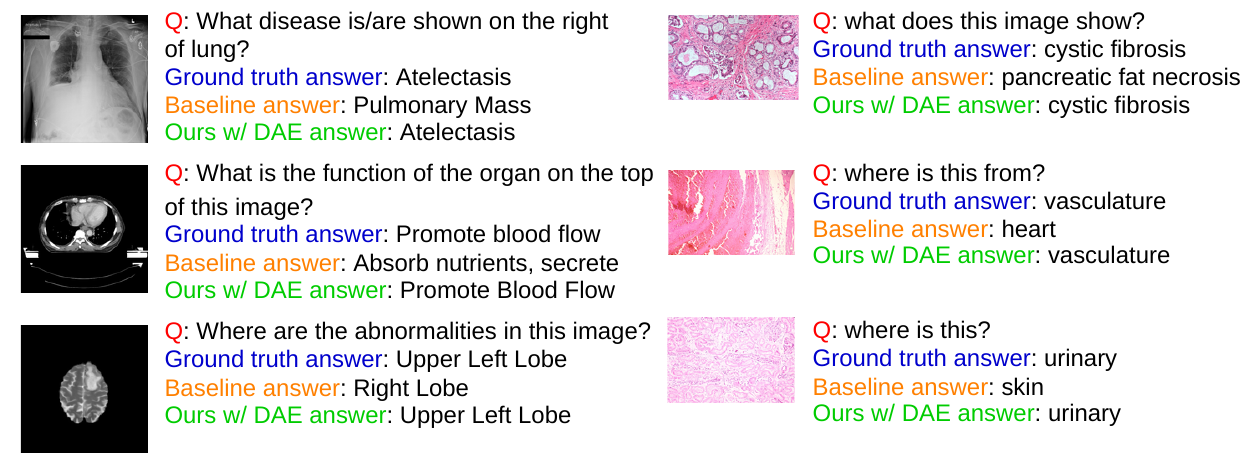}
\caption{Qualitative examples comparing the baseline and proposed DAE-based models. The left column shows examples from SLAKE, while the right column shows examples from PathVQA.}
\label{fig:qualitative_examples}
\end{figure}

Fig.~\ref{fig:qualitative_examples} presents qualitative examples in which the proposed DAE-based model predicts the ground-truth answers correctly, whereas the baseline model produces incorrect responses. The left column shows examples from SLAKE, while the right column shows examples from PathVQA.

Overall, these results show that the DAE-based approach is more robust than both direct projection and standard AE-based representation learning. While baseline and AE models experience larger performance drops as noise increases, the DAE-based models maintain higher accuracy under noisy embedding conditions. This suggests that denoising representation learning helps produce visual embeddings that are less sensitive to perturbations before being projected into the LLM embedding space.

\subsection{Limitations and Future Work}

The denoising autoencoder is trained using Gaussian corruption applied to visual embeddings produced by the pretrained vision encoder. While this setting improves robustness against synthetic perturbations, it may not fully represent the diverse noise characteristics and artefacts found across real medical imaging modalities and acquisition settings. Future work may investigate more realistic corruption strategies, modality-specific noise simulation, or adaptive denoising objectives.

The current framework also employs a relatively simple visual mapping strategy to project visual representations into the language model input space. More advanced approaches for mapping visual representations into the LLM embedding space, such as query-based transformers, cross-attention mechanisms, or knowledge-guided adapters, may further improve visual-language interaction and answer generation performance.

Finally, the current study focuses on a frozen vision encoder and parameter-efficient adaptation of the language model. Future work may explore hybrid fine-tuning strategies, larger-scale medical multimodal pretraining, and integration with external medical knowledge to further enhance open-ended medical reasoning capabilities.

\section{Conclusion}
This paper presented a noise-aware Med-VQA framework that incorporates a denoising autoencoder before the visual mapping stage, enabling visual embeddings from a frozen vision encoder to be denoised before being projected into the input space of an LLM. The proposed two-stage training strategy enables the model to learn robust and noise-aware visual representations before they are projected into the LLM embedding space for answer generation. Experimental results across multiple Med-VQA benchmark datasets show that the proposed approach provides competitive clean performance while consistently enhancing robustness under noisy embedding conditions, suggesting that denoising-based representation learning can produce more stable visual embeddings for downstream Med-VQA tasks.

\bibliographystyle{splncs04}
\bibliography{z-references}
\end{document}